\def\year{2021}\relax
\pdfoutput=1
\documentclass[letterpaper]{article} % DO NOT CHANGE THIS
\usepackage{aaai21}  % DO NOT CHANGE THIS
\usepackage{times}  % DO NOT CHANGE THIS
\usepackage{helvet} % DO NOT CHANGE THIS
\usepackage{courier}  % DO NOT CHANGE THIS
\usepackage[hyphens]{url}  % DO NOT CHANGE THIS
\usepackage{graphicx} % DO NOT CHANGE THIS
\usepackage{natbib}  % DO NOT CHANGE THIS AND DO NOT ADD ANY OPTIONS TO IT
\usepackage{caption} % DO NOT CHANGE THIS AND DO NOT ADD ANY OPTIONS TO IT
\usepackage{amsmath,amsfonts,amssymb}
\usepackage[switch]{lineno}
\usepackage{color}
\usepackage{tabularx}
\usepackage{booktabs} % For formal tables
\usepackage{subfigure}
\usepackage{xspace}
\usepackage{multirow}
\newcommand{\eat}[1]{}

\newcommand{\kw}[1]{\textit{#1}\xspace}

\newcommand{\umean}{\kw{UserMean}}
\newcommand{\imean}{\kw{ItemMean}}

\newcommand{\svd}{\kw{SVD++}}

\newcommand{\soreg}{\kw{SocialReg}}

\newcommand{\rste}{\kw{RSTE}}
\newcommand{\graphrec}{\kw{GraphRec}}
\newcommand{\danser}{\kw{DANSER}}
\newcommand{\our}{\kw{ASR}}
\newcommand{\ouru}{\kw{ASR-U}}
\newcommand{\ouri}{\kw{ASR-I}}

\newcommand{\ourgats}{\kw{ASR-w/-$G_S$-GAT}}
\newcommand{\ourgatr}{\kw{ASR-w/-$G_R$-GAT}}

\title{Attentive Social Recommendation: Towards User And Item Diversities}

\author{%
Dongsheng Luo$^1$ \quad Yuchen Bian$^2$ \quad Xiang Zhang$^1$ \quad Jun Huan$^3$\\}

\affiliations{
    $^1$The Pennsylvania State University, $^2$Baidu Research, USA, $^3$StylingAI Inc.\\
    $^1$\{dul262,xzz89\}@psu.edu, $^2$yuchenbian@baidu.com, $^3$lukehuan@shenshangtech.com
}

\begin{document}

% \clearpage
\maketitle

% \linenumbers

\begin{abstract}
Social recommendation system is to predict unobserved user-item rating values by taking advantage of user-user social relation and user-item ratings. However, user/item diversities in social recommendations are not well utilized in the literature. Especially, inter-factor (social and rating factors) relations and distinct rating values need taking into more consideration. In this paper, we propose an attentive social recommendation system (ASR) to address this issue from two aspects. First, in ASR, Rec-conv graph network layers are proposed to extract the social factor, user-rating and item-rated factors and then automatically assign contribution weights to aggregate these factors into the user/item embedding vectors. Second, a disentangling strategy is applied for diverse rating values. Extensive experiments on benchmarks demonstrate the effectiveness and advantages of our ASR.

% First, social influences and rating factors  

% of social influences and rating factors should be dynamically considered with distinct attention contributions for different users and items. This means that (1) users may be more affected by their social relations or their own rating histories; (2) ratings for an item may include both users' subjective and item's objective effects. However, existing works overlook the inter-factor relations and the diversity of rating propagation. In this paper, we propose a novel intra-factor and inter-factor attentive social recommendation system (ASR). ASR consists of stacked attentive Rec-conv network layers to flexibly differentiate multiple factors for user/item representation learning. In each Rec-conv layer, a disentangling strategy is applied for diverse ratings. Then the attention mechanism aggregates multiple factors into user/item embedding. 
% % ASR can also alleviate the over-smoothing issue when multiple Rec-conv layers are stacked. 
% Extensive experiments on two benchmarks demonstrate the effectiveness and advantages of our ASR.

\end{abstract}

\section{Introduction}
\label{sec:intro}

Recommendation system aims to predict unobserved ratings based on users' historical purchases. Users are also involved in social relations where they often acquire and propagate preferences. \textit{Social recommendation} is to leverage the social factor in recommendation systems. It has been verified to be effective for alleviating the data sparsity and cold-start issue existing in traditional collaborative filtering-based recommendation methods~\citep{wu2019dual,fan2019graph}.

\begin{figure}[h]
    \centering
    \includegraphics[width=3.2in]{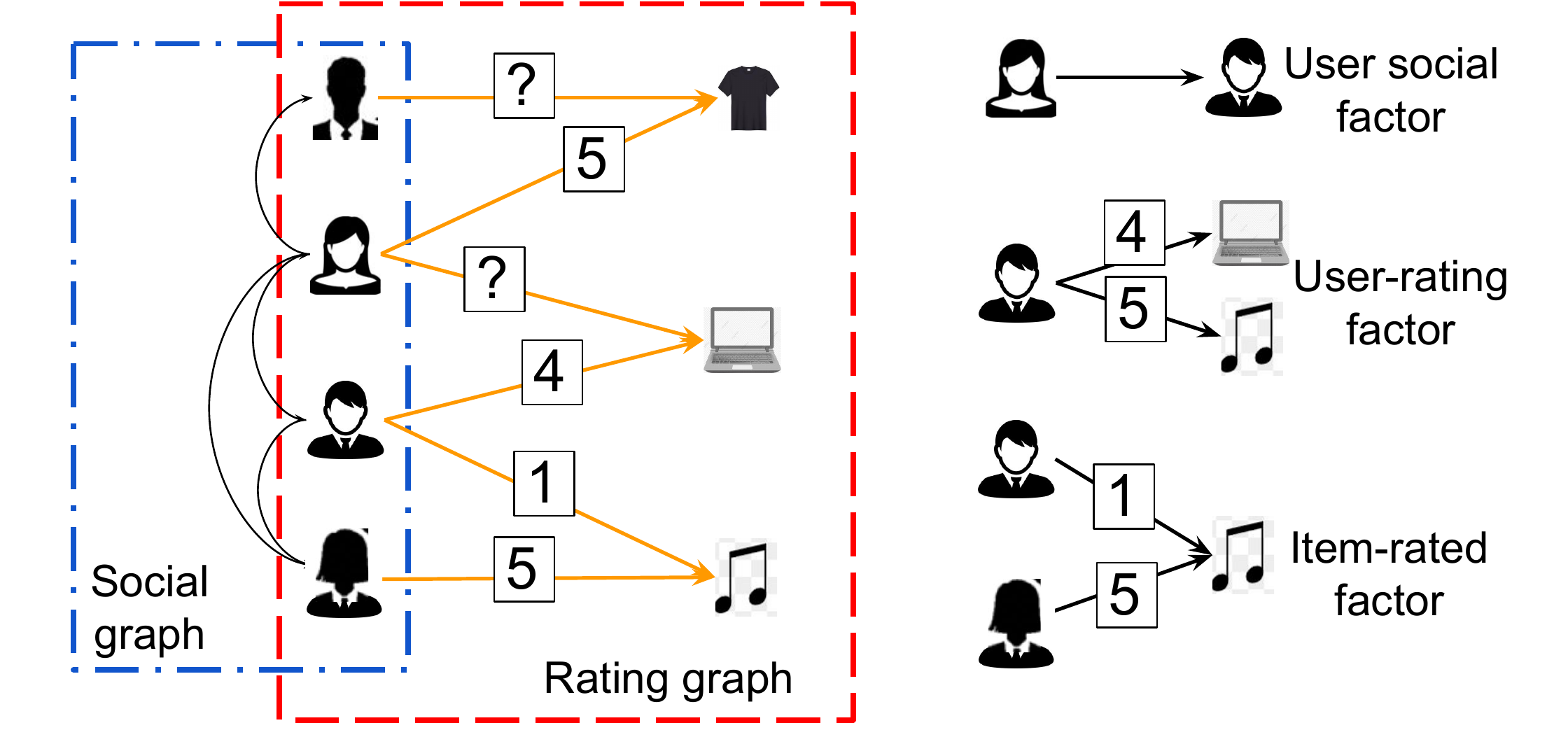}
    \vspace{-0.05in}
    \caption{Predict the unobserved ratings in social recommendation by considering user social factor, user-rating factor, and item-rated factor.}
    \label{fig:input}
    \vspace{-0.1in}
\end{figure}

In social recommendation, both user-user social relations and user-item rating data are often represented by graph structures. As shown in Figure~\ref{fig:input}, we structure the user-user relationship as a graph (blue part) and user-item ratings as a bipartite graph (red part) with edge weights representing rating values. Usually, rating values range from ``1'' to ``5'' with ``1'' as dislike and ``5'' as like most, for instance, in the two benchmark datasets in the experiments Ciao and Epinions. Taking advantage of graph neural network (GNN), e.g., graph convolutional network (GCN)~\citep{gcn} and graph attention network (GAT)~\citep{gat}, recent works are proposed to extract features from the social graph and the rating graph~\citep{wu2019dual,fan2019deep,fan2019graph}.

\begin{figure}[!t]
    \centering
    \includegraphics[width=3.0in]{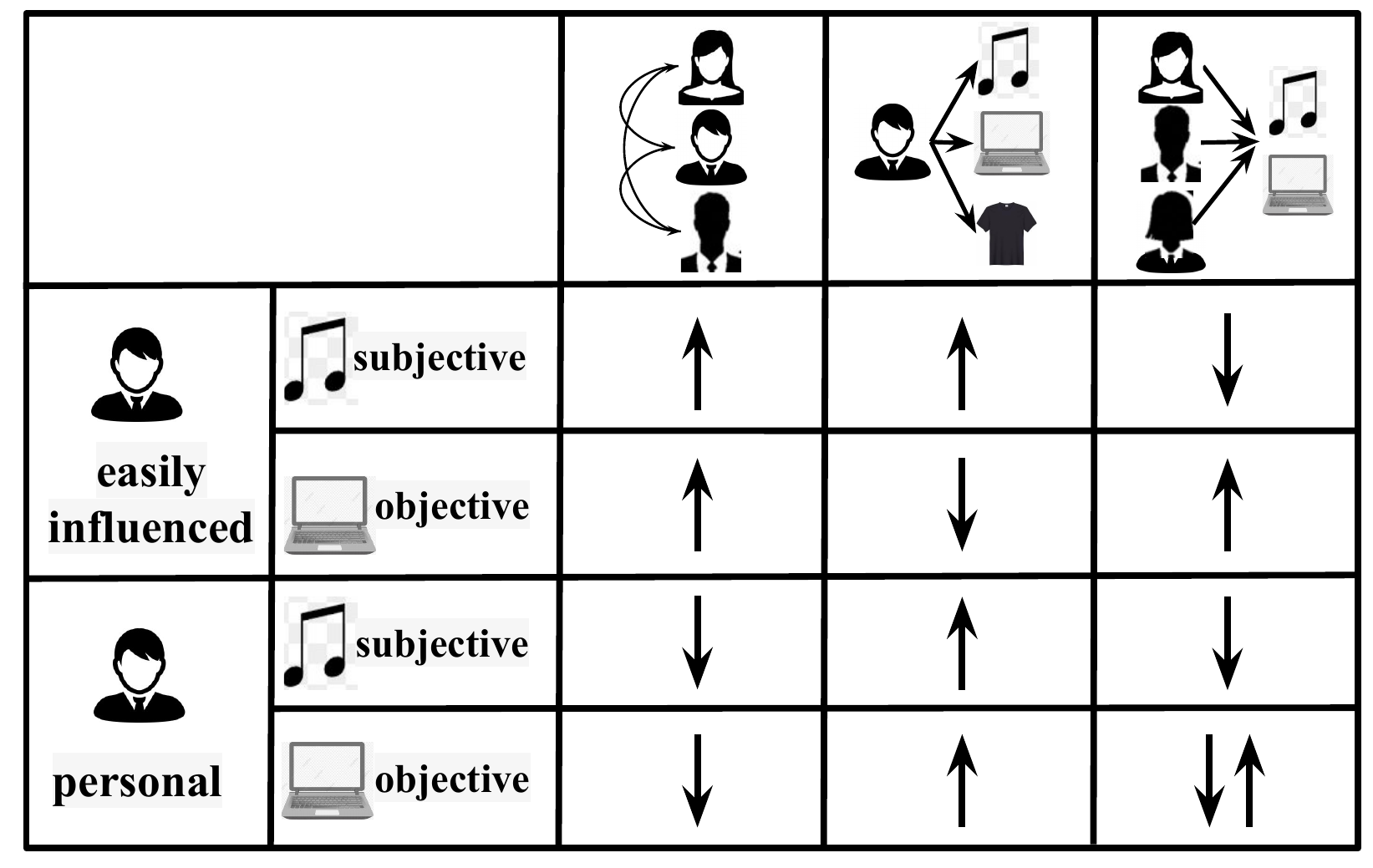}
    \vspace{-0.05in}
    \caption{Motivation of our attentive social recommendation: considering inter-factor contributions for user/item diversity. $\uparrow/\downarrow$ means more/less attention weights.}
    \label{fig:motivation}
    \vspace{-0.1in}
\end{figure}

% \textcolor{red}{definition of factors not clear} 
As demonstrated in Figure~\ref{fig:input}, three factors are often taken into account in social recommendation. \textit{Social factor} reflects that a user's ratings may be influenced by the neighbors in the social graph. In the rating graph, we define \textit{user-rating factor} as the effect on an individual user from all her/his ratings and \textit{item-rated factor} as the impact on an item from all its ratings.

Despite the successes of feature extractors, diversity of users and items are not well investigated in the literature. There still are two main challenges. 

First, existing works overlook the different contributions of the multiple factors when considering user/item diversities. Shown in Figure \ref{fig:motivation}, we take four cases for illustration. For users, some users (the first two rows) are easily influenced by friends, i.e., more social factor should be paid attention; while some (the last two rows) have their unique preferences reflected by their own ratings (more user-rating factor and less social factor). Moreover, evaluating an item could be more subjective or objective. We should recommend a song (subjective items) based on users' different tastes which means that more user-rating factor needs considering. But recommending a computer (objective items) would require more its overall ratings, i.e., more item-rated factor. User/item diversities in real scenarios are more complex than the four exemplars. Thus attentive inter-factor contributions should be emphasized. 
% In the real scenario, attentions among multiple factors should be emphasized. 
However, existing methods only separately extract some factor features. For example, \graphrec~\citep{fan2019graph} separately applies a GAT in either the social graph or the rating graph and then simply concatenates the extracted features for further rating prediction. 
% One reason for their lack of capabilities to capture inter-factor features is that the intra-factor GNN layers cannot be deeply stacked among factors. But it is shown that the depth of a GNN is critical to its learning capacity~\citep{loukas2020what}. 

Second, distinct rating values are not well exploited. ``like'' and ``dislike'' ratings may propagate in social and rating graphs in different patterns. But existing work, for example, \danser~\citep{wu2019dual} does not distinguish the edge weights (i.e., rating values) in the rating graph and only utilizes rating values in the loss computation. 

To tackle the two challenges, in this paper, we propose an  \emph{\underline{A}ttentive \underline{S}ocial \underline{R}ecommendation (\our)} model to attentively fuse multiple factors for user/item diversities in social recommendation.

% Intuitively, we propose an stackable Rec-conv layer to flexibly assign contribution weights to in multiple factors. s

For the first challenge, in ASR, a new graph neural network architecture, Rec-conv layer, is proposed. In each Rec-conv layer, GNNs are applied to extract the three aforementioned factors from the social graph and the rating graph. Attention mechanisms are utilized as well in each Rec-conv layer to automatically assign contribution weights on the three factors and to obtain factor-fused user/item embedding vectors.
 
% The Rec-conv layer can be deeply stacked to effectively extract and aggregate multiple intra-factor and inter-factor features from both social graph and user-item rating graph.

% To update each user embedding, the Rec-conv layer extracts and combines three intra-factors: (i) the user vector in the previous Rec-conv layer, which demonstrates the user's own profile; (ii) social factor, describing friends’ influence from the social graph; and (iii) user-rating factor, reflecting user’s rating history. Then an attention mechanism is followed to assign contribution weights on these three factors to obtain an inter-factored user embedding. Similarly, an updated item embedding attentively fuses (i) the previous item vector and (ii) the item-rated factor in the rating graph, which describes its overall rating. 

% *******Note that both the user-rating and item-rated information are extracted by GNNs from the rating graph. So an item embedding also identifies users subjective and items objective effects. In this sense, ASR can fuse intra-factor and inter-factor features by the deeply stacked Rec-conv layers.

% are followed by fully connected layers (a.k.a., MLP) to get a supervised loss to guide training~\citep{he2017neural}. 
% \textcolor{red}{cite who use MLP?}. 
For the second challenge, ASR adopts a disentangling strategy to distinguish the propagation of ``like'' and ``dislike'' ratings. Specifically, for each rating value, we first induce a subgraph from the entire rating graph. Then we combine all the GNN-extracted user-rating and item-rated factors from each subgraph into the user/item embedding in each Rec-conv layer.
% Experimental results verify the advance of this disentangled strategy (Section \ref{sec:disentabgled}).

% Moreover, technically, the over-smoothing issue exists when stacking multiple GNN layers~\citep{li2018deeper}. But the inter-factor attention mechanism of the Rec-conv layers in ASR can alleviate the over-smoothing by actively selecting useful information to pass forward to the next layer. 
% Detailed discussion is provided in Section \ref{sec:over-smoothing}.
% \textcolor{red}{Besides, it can also relieve the vanishing gradient problem which comes with a standard deep neural graph. Is there any evaluation? If no, then delete}. 

Extensive experimental results on two real-world datasets verify the better effectiveness and efficiency of ASR than state-of-the-art methods. 
% \textcolor{red}{ACN outperform by (2.98\%, and 4.05\%) on MAE and (0.92\%, 1.05\%) on RMSE in Ciao and Epinions dataset, respectively.}
{We also conduct ablation study to demonstrate the effectiveness of the inter-factor attention mechanism, the disentangling strategy and GNNs, and to examine the sensitivity of the stacked the Rec-conv layers to the over-smoothing issue~\citep{li2018deeper}.}% \% and \%

Our major contributions are summarized as follows.
\begin{itemize}
    % \item The intrinsic quality of user and item in social recommendations are investigated in this paper, which is overlooked in previous studies.
    \item Diversities of users and items are investigated. Inter-factor contributions and distinct rating propagation are vital to social recommendation. 
    \item A novel attentive social recommendation (ASR) system with stacked Rec-conv layers is proposed to effectively fuse multiple factors for user/item representation learning.
    \item Extensive experiments on two benchmarks demonstrate the advantages of ASR both in effectiveness and efficiency.
\end{itemize}

\section{Related Work}\label{sec:relatedwork}
% Recommendation systems provide personalized products to users based on the rating data. 
\textbf{Classic recommendation system.} Collaborating filtering based methods are widely used in recommendation systems~\citep{koren2010factor,koren2009matrix,he2017neural}. Most methods model a user's preference by collecting and analyzing rating information from other users with matrix factorization techniques. Recently, deep neural networks have been applied in this task~\citep{he2017neural,guo2017deepfm,pinsage,wu2019feature,krishnan2019modular,fan2019deep,fan2019deepijcai,jin2020multi,lei2020social}. An overview can be found in~\citet{zhang2019deep}.

% Moreover, 
\textbf{Social recommendation.} Investigating user social relationship in recommendation has drawn increasing attention because of its capability to alleviate the data sparsity and cold-start problem~\citep{tang2013social,ma2011recommender,fan2019graph}. 
% In ~\citeauthor{ma2008sorec}~\citeyear{ma2008sorec}, the authors proposed a probabilistic matrix factorization based method to solve the data sparsity and poor prediction accuracy problems. 
For instance, \soreg~\citep{ma2011recommender} is a matrix factorization method with social regularization. \rste~\citep{ma2009learning} fuses the users' tastes and their friends influences together with a probabilistic 
framework. TrustMF~\citep{yang2016social} tries to capture users’ reciprocal influence to learn low-dimensional user embedding in truster space and turstee space. 
SoDimRec~\citep{tang2016recommendation} investigates the heterogeneous relations and weak dependency connections in social graphs. 
DiffNet~\citep{wu2019neural} introduces an influence propagation mechanism to stimulate the recursive social diffusion process in social recommendation. Attention mechanisms are introduced in DiffNet++~\citep{wu2020diffnet}, EATNN~\citep{chen2019efficient}, DGRec~\citep{song2019session} and SoRecGAT~\citep{vijaikumar2019sorecgat}. 
A survey of social recommendation can be found in~\citet{tang2013social}.

\textbf{GNN-based social recommendation.} More recently, GNN~\citep{gcn,gat} has been used in social recommendation due to its abilities of aggregating local neighbors information in graphs~\citep{monti2017geometric,pinsage,wu2019dual,fan2019graph,yu2020enhance}. 
In~\citet{monti2017geometric}, the authors generalize GNN to multiple graphs and to learn user/item representations. 
GraphRec~\citep{fan2019graph} and DSCF~\citep{fan2019deep} apply GATs~\citep{gat} in the user-user social graph and user-item rating graph separately to extract user/item features. DANSER~\citep{wu2019dual} adopts GAT to learn user/item static and dynamic embedding vectors. 

% DGRec~\citep{song2019session} utilizes RNN to capture the dynamic user behaviors and an attention network to model the context dependent social influence.

Nevertheless, they cannot effectively and attentively fuse social and rating factors and distinct ratings for user/item diversities in social recommendations.

% Nevertheless, they only learn various factor features from either social or rating graph. They cannot effectively and attentively fuse social and rating factors. They also lack capabilities to stack into deep models and to effectively handle various rating values. On the other hand, the deeply stacked Rec-conv layers in our proposed ASR can effectively and attentively fuse the intra-factor and inter-factor features into user/item representations.

% However, in real-world recommendation systems, users have diverse habits. Some users are influenced more by friends' opinions, while others have their unique preferences. Different influences of these high-level factors (influence from friends and influence from purchase histories) are overlooked in GraphRec and DANSER.
% Instead, they differentiate the low-level influence inside each network. For instance, for a user, attention networks used in these two methods are to differentiate influences from different friends.

% In summary, existing methods cannot flexibly pay attention to social relationship and user-item rating information. They overlook diversity of rating  patterns for users and items which is the main focus of this paper.

\begin{table}[!ht]
    \centering
    \begin{small}
    \caption{Main symbols}
    \label{tab:symbols}
    \begin{tabular}{c|l}
    \hline
    % Symbol &Definition\\
    \multicolumn{1}{c|}{\textbf{Symbol}} & \multicolumn{1}{c}{\textbf{Definition}}\\
    \hline
	    $\mathcal{U}$ & the user set; $N=|\mathcal{U}|$\\
	    $\mathcal{I}$ & the item set; $M=|\mathcal{I}|$\\
	    $\mathcal{R}$ & the set of observed user-item rating pairs\\
	    $G_S$ & the social graph $G_S=(\mathcal{U}, \mathcal{S})$ with $\mathcal{S}$ as the edge set\\
	    $G_R$ & the user-item rating bipartite graph $G_R=(\mathcal{U}, \mathcal{I},\mathcal{R})$\\
        $\mathbf{S}$ & the binary adjacency matrix of social graph $G_S$\\
        $\mathbf{R}$ & the integer-rating matrix of $G_R$ $(r_{xy}\in \{1\cdots K\})$\\
        $\mathbf{U}^{(l)}$ & the user repre. matrix of the $l$-th Rec-conv layer\\
        $\mathbf{U}^{(l,\mathcal{U}\rightarrow u)}$ & the social factor repre. matrix of the $l$-th layer\\
        $\mathbf{U}^{(l,\mathcal{I}\rightarrow u)}$ & the user-rating factor repre. matrix of the $l$-th layer\\
        $\mathbf{I}^{(l)}$ &the item repre. matrix of the $l$-th layer\\
        $\mathbf{I}^{(l,\mathcal{U}\rightarrow i)}$ & the item-rated factor repre. matrix of the $l$-th layer \\
    \bottomrule
    \end{tabular}
    % \vspace{-0.2in}
    \end{small}
\end{table}

\section{Attentive Social Recommendation}\label{sec-3:framework}
In this section, we introduce the framework of Attentive Social Recommendation which can dynamically extract and fuse social, user-rating and item-rated factors via stacking the newly proposed Rec-conv layers. 
% We start from preliminaries.

\subsection{Preliminaries}\label{sec:problem}
In this paper, we represent user/item set as $\mathcal{U}$ ($N=|\mathcal{U}|$) and $\mathcal{I}$ ($M=|\mathcal{I}|$), respectively. A social graph $G_{S}=(\mathcal{U},\mathcal{S})$ and a user-item rating bipartite graph $G_{R}=(\mathcal{U}, \mathcal{I}, \mathcal{R})$ are given. In $G_{S}$, $\mathcal{S}$ describes users social connections. Let its adjacency matrix be $\mathbf{S}\in \{0,1\}^{N\times N}$ with 1/0 representing if social relation exists between two users or not. In $G_{R}$, $\mathcal{R}$ records the observed user-item rating pairs. Let $\mathbf{R} \in \mathbb{R}^{N\times M}$ record user-item rating values. A user-item pair $(x,y)\in \mathcal{R}$ $(x\in \mathcal{U}, y\in \mathcal{I})$ is assigned with an existing integer rating value $r_{xy}\in \mathbf{R}$ which ranges from 1 (``like least'') to $K$ (``like most''). We also initialize the unobserved ratings in $\mathbf{R}$ with 0s. The goal of social recommendation is to estimate the unobserved ratings.  
% All rating values are stored in the adjacent matrix .  
% The goal of social recommendation is to estimate the unobserved rating value $r_{ui}$ for the unconnected user-item pairs. 
Main notations of this paper are also summarized in Table~\ref{tab:symbols}.

\begin{figure*}[h!]
    \centering
    \subfigure[The framework]{\includegraphics[width=3.0in]{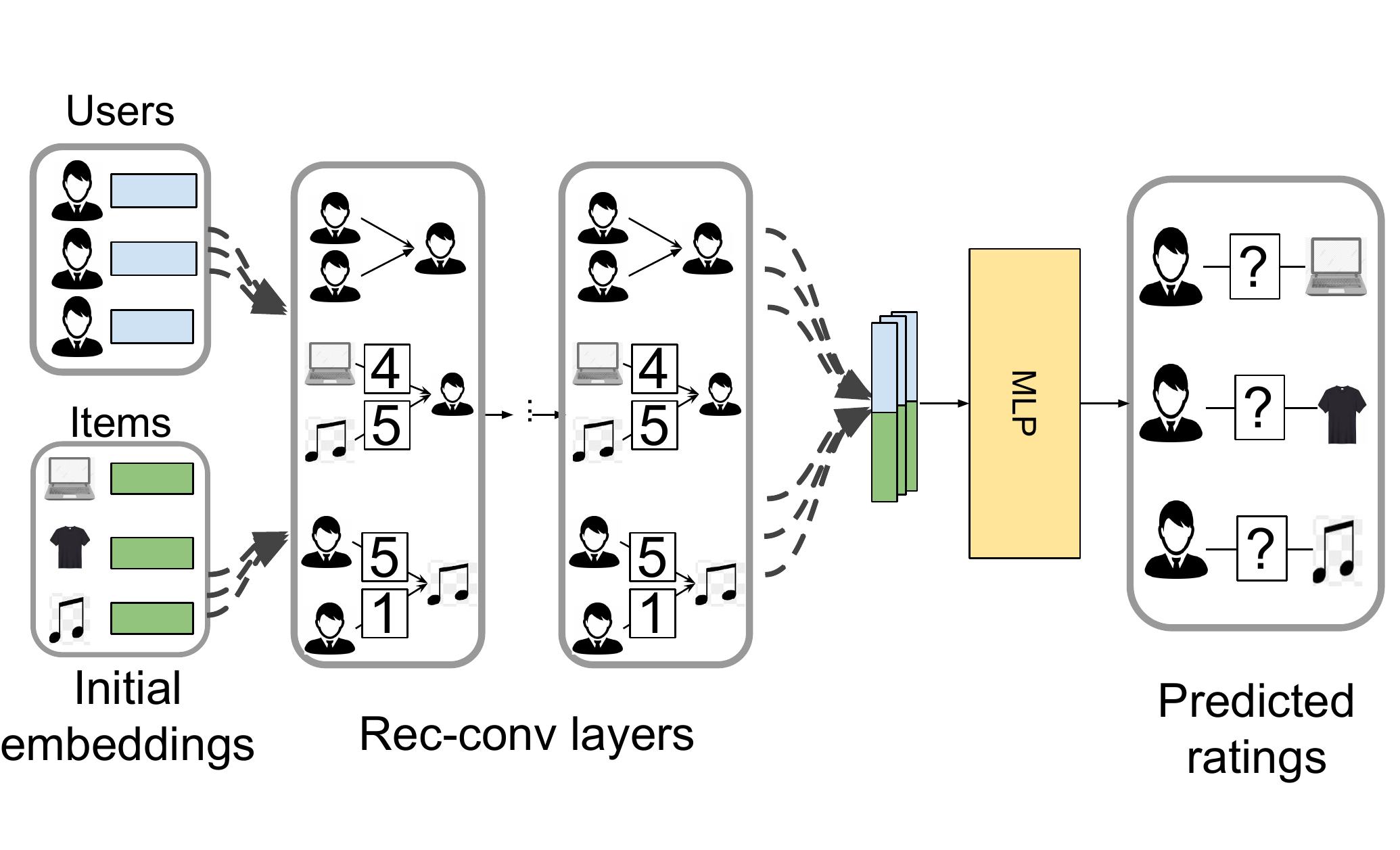}\label{fig:arch}}\hspace{2em}
    \subfigure[The $l$\textsuperscript{th} Rec-conv layer]{\includegraphics[width=3.2in]{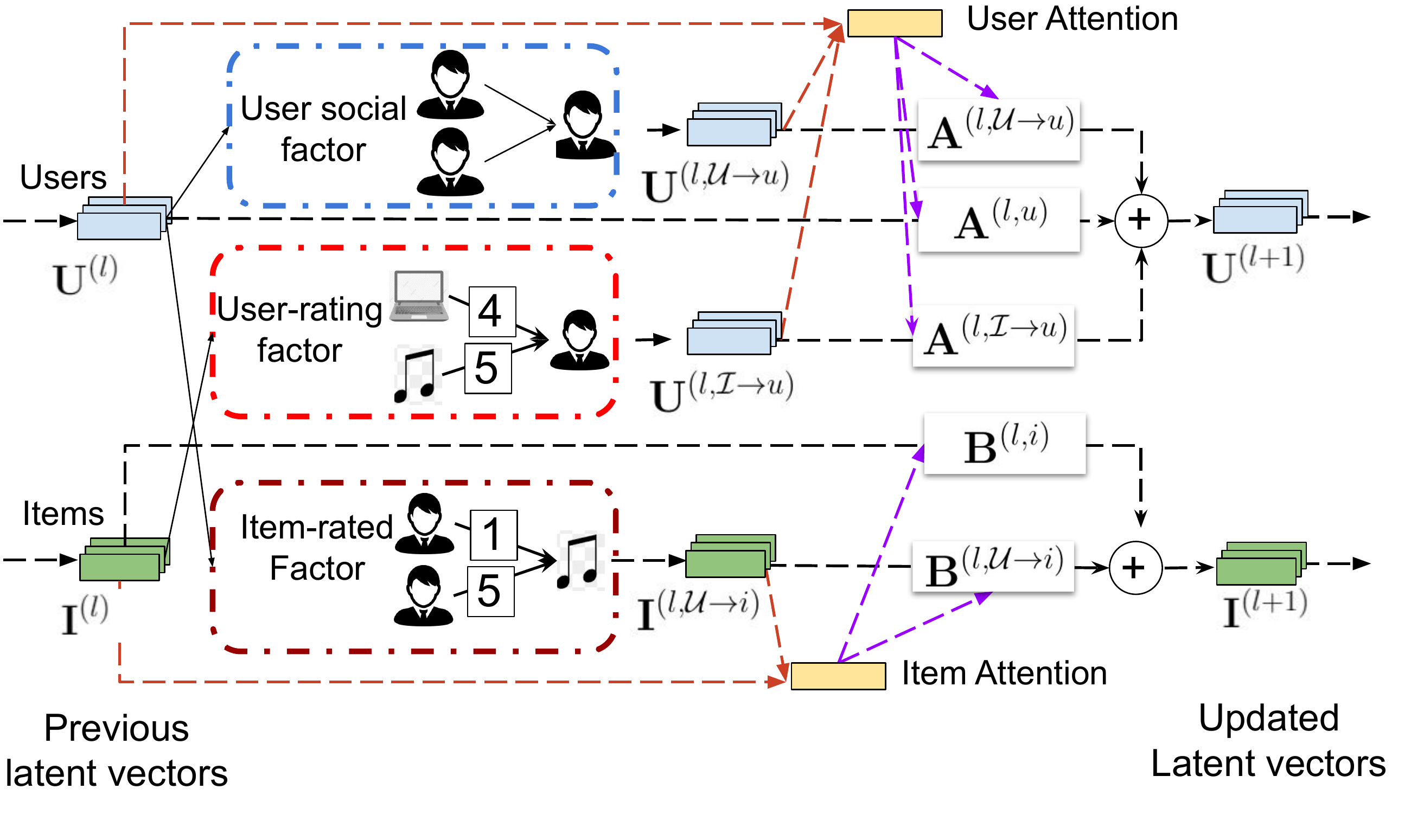}\label{fig:layer}}
    % \vspace{-0.1in}
    \caption{(a) The framework of ASR. Blue/green bars denote user/item latent vectors. (b) Social factor $\mathbf{U}^{(l,\mathcal{U}\rightarrow u)}$, user-rating factor $\mathbf{U}^{(l,\mathcal{I}\rightarrow u)}$ and item-rated factor $\mathbf{I}^{(l,\mathcal{U}\rightarrow i)}$ are extracted and attentively fused in the user/item embedding updates.}
    % Note that the user and item embedding are shared in the bipartite rating graph. So they are also fused in the user-rating factor and the item-rated factor.}
    % also fused in each Rec-conv layer since both the user-rating and item-rated aggregations are from the bipartite rating graph.
    \label{fig:framework}
    % \vspace{-0.1in}
\end{figure*}

\subsection{The Framework}\label{sec:method}
ASR targets to attentively fuse three factors (user-user social factor, user-rating factor, and item-rated factor) into user and item embedding. The overall architecture of ASR is shown in Figure~\ref{fig:arch}.

% extract social factor features from the social graph $G_S$, user-rating factor and item-rated factor features from the bipartite rating graph $G_R$ and then

% \begin{figure}[t]
%     \centering
%     \includegraphics[width=3.3in]{Figure/framework_Recconv_full3.pdf}
%     \vspace{-0.3in}
%     \caption{The framework. Blue and green rectangles denote user and item vectors, respectively.}
%     \label{fig:arch}
% \end{figure}

We represent users and items with $D$-dimension latent vectors as $\mathbf{U} \in \mathbb{R}^{N\times D}$ and $\mathbf{I} \in \mathbb{R}^{M\times D}$,  respectively. 
% The $x$\textsuperscript{th} row $\mathbf{u}_x$ in $\mathbf{U}$ is the embedding of user $x$, and the $y$\textsuperscript{th} row $\mathbf{i}_y$ in $\mathbf{I}$ is the embedding of item $y$. 
User and item embedding vectors are updated by forwarding through stacked Rec-conv neural network layers. 
% Inputs of a Rec-conv layer are user and item vectors from the previous layer, and outputs are updated vectors. 

% The Rec-conv layer aims to dynamically and attentively fuse multiple factors with distinct contribution weights into the users and items embedding. 
%  should be dynamically assigned to different information sources for different users/items.
In each Rec-conv layer, the social factor features are extracted from the social graph $G_S$; user-rating factor and item-rated factor features are obtained from the bipartite rating graph $G_R$. In the meanwhile, the user and item embedding vectors are updated by attentively fusing all the three factors.

% Intuitively, if a user is more easily affected by friends, then social factor should be assigned with more weights. If a user has a unique taste, the user-rating factor should be emphasized. For a more subjective item (e.g., a song), the user tastes (i.e., user-rating factor) may need more attention. In contrast, for a more objective item (e.g., a computer) the item overall rating (i.e., item-rated factor) will help the recommendation. After passing through deeply stacked Rec-conv layers, the user and item vectors are updated by attentively fusing all factors.

% Attentive contribution weights in each Rec-conv layer are automatically updated in the back propagation. 
% Technically, we design the following process and a learning objective.

% \subsection{Learning Objective}\label{sec:alg}
After obtaining the final user/item embedding, we follow the common setting in the literature~\citep{wu2019dual,fan2019graph} to predict the unobserved rating value for a user $x$ on an item $y$. We concatenate the user and item vectors and feed them 
% , i.e. $\mathbf{u}_x^{(L)}$ and $\mathbf{i}_y^{(L)}$. We then feed the concatenation of $\mathbf{u}_x^{(L)}$ and  $\mathbf{i}_y^{(L)}$ 
 into an MLP to predict the rating value $r'_{xy}$. During training, given the ground-truth rating $r_{xy}$, we take the mean square error as the training objective:
\begin{equation}
    \label{eq:objective}
    Loss = \frac{1}{2|\mathcal{R'}|}\sum_{(x,y)\in \mathcal{R'}}(r'_{xy}-r_{xy})^2
\end{equation}
where $\mathcal{R'}\subset\mathcal{R}$ is the training set containing observed  ratings.

% \begin{figure}[t]
%     \centering
%     \includegraphics[width=3.5in]{Figure/Recconv.pdf}
%     \vspace{-0.3in}
%     \caption{The $l$\textsuperscript{th} Rec-conv layer}
%     \label{fig:layer}
% \end{figure}

\section{The Rec-conv Layer}\label{sec:reconv}
In this section, we introduce our Rec-conv layer to attentively aggregate multiple factors from social and rating graphs in the user/item embedding. A diagram is shown in Figure~\ref{fig:layer}. Two main processes in the Rec-conv layer are to update user embedding (Section \ref{sec:user_emb}) and item embedding {Section \ref{sec:item_emb}}.

\subsection{Update User Embedding $\mathbf{U}^{(l)}\rightarrow\mathbf{U}^{(l+1)}$}\label{sec:user_emb}
Users are influenced by various factors, such as their own tastes (user-rating factor) and social effects from friends (social factor). To model these factors, in the $l$\textsuperscript{th} Rec-conv layer, we use GNNs to generate two kinds of user latent vectors: \textit{user social vectors} $\mathbf{U}^{(l,\mathcal{U}\rightarrow u)}$ and \textit{user-rating vectors} $\mathbf{U}^{(l,\mathcal{I}\rightarrow u)}$. An attention mechanism is used to attentively aggregate all factors. In the following, we take GCN~\citep{gcn} as an example. Note that Other GNNs can also be used, we evaluate their performance in {Section \ref{sec:disentabgled}}.

\subsubsection{User Social Vector $\mathbf{U}^{(l,\mathcal{U}\rightarrow u)}$} 
Social vectors are extracted by applying a GCN to propagate
% and aggregate 
local neighbor's information in the social graph. Formally:
\begin{equation}
    % \mathbf{U}^{(l,\mathcal{U}\rightarrow u)}=\sigma ({\tilde{\mathbf{D}}_S}^{-\frac{1}{2}}\tilde{\mathbf{S}}{\tilde{\mathbf{D}}_S}^{-\frac{1}{2}}\mathbf{U}^{(l)}\mathbf{W}^{(l)}_S)
    \mathbf{U}^{(l,\mathcal{U}\rightarrow u)}=\sigma (\tilde{\mathbf{S}}\mathbf{U}^{(l)}\mathbf{W}^{(l)}_S)
\end{equation}

Here, $\mathbf{U}^{(l)} \in \mathbb{R}^{N\times D}$ is the user latent matrix of the $l$\textsuperscript{th} layer. $\mathbf{U}^{(l,\mathcal{U}\rightarrow u)}$ is the user social factor we aim to extract. $\tilde{\mathbf{S}} = {{\mathbf{D}}_S}^{-\frac{1}{2}}(\mathbf{S} + \mathbf{I}_N){{\mathbf{D}}_S}^{-\frac{1}{2}}$ is the modified transition matrix of $G_S$ by adding self-connections ($\mathbf{I}_N$ is the identity matrix) and being normalized by the node-degree diagonal matrix $\mathbf{D}_S$, which is a standard formalization in GCN ~\citep{gcn}. In such way, $\tilde{\mathbf{S}}\mathbf{U}^{(l)}$ can aggregate user neighbors influences into the target social factor. 

$\mathbf{W}^{(l)}_S\in \mathbb{R}^{D\times D}$ is a layer-specific trainable weight matrix. $\sigma(\cdot)$ denotes an \textcolor{black}{activation function, e.g., ReLU}.  And $\mathbf{U}^{(0)}$ can be \textcolor{black}{randomly initialized or with user profile matrix if available.}

% $\tilde{\mathbf{S}} = {{\mathbf{D}}_S}^{-\frac{1}{2}}(\mathbf{S} + \mathbf{I}_N){{\mathbf{D}}_S}^{-\frac{1}{2}}$ is the adjacency matrix of $G_S$ with self-connections. $\mathbf{I}_N$ is the identity matrix. $\tilde{\mathbf{D}}_S$ is the diagonal matrix with its diagonal entries $\tilde{d}_{xx} = \sum_y \tilde{s}_{xy}$ and $\mathbf{W}^{(l)}_S\in \mathbb{R}^{D\times D}$ is a layer-specific trainable weight matrix for aggregating neighbor users influences. $\sigma(\cdot)$ denotes an \textcolor{black}{activation function, e.g., ReLU}.  And $\mathbf{U}^{(0)}$ can be \textcolor{black}{randomly initialized or with user profile matrix if available.}

\subsubsection{User-Rating Vector $\mathbf{U}^{(l,\mathcal{I}\rightarrow u)}$}\label{sec:user-rateing-vector}
The user-rating vector for a user reflects the user's preference. And it is extracted from all the ratings in the bipartite rating graph. However, a GCN on the entire rating graph $G_R$ will mix distinct rating values which may represent  opposite attitudes (e.g., value 1 and 5). So we apply a disentangling strategy to extract diverse rating effects. Note that we also verify the effectiveness of this disentangling strategy in Section \ref{sec:disentabgled}.

Specifically, we first induce rating subgraphs based on the $K$ diverse rating values. We let $\mathcal{R}_k = \{(x\in \mathcal{U}, y\in \mathcal{I})| r_{xy}=k\}$ be the subset of rating-pair with rating $k\in \{1\cdots K\}$ and $\mathbf{R}_k$ be the corresponding rating matrix. Thus, $\mathcal{R} = $ $\cup_{k=1}^{K}\mathcal{R}_k$, and  $\mathcal{R}_i \cap \mathcal{R}_j = \emptyset$ for $i\neq j$. 
% Specifically, as a pre-processing step, according to rating values, we divide the observed user-item ratings $\mathcal{R}$ into several disjointed subsets $\mathcal{R}_1, \mathcal{R}_2 \cdots \mathcal{R}_K$, where $\mathcal{R}_k = \{r_{xy}|x\in \mathcal{U}, y\in \mathcal{I}, r_{xy}=k\}$. Thus, $\mathcal{R} = $ $\mathcal{R}_1 \cup \mathcal{R}_2 \cdots \cup \mathcal{R}_K$, and for $i\neq j$, $\mathcal{R}_i \cap \mathcal{R}_j = \emptyset$. The edge-induced subgraph of $\mathcal{R}_k$ is denoted by $G_{Rk}$ and the corresponding adjacency matrix is $\mathbf{R}_k \in \mathbb{R}^{N\times M}$. 
Then $K$ channels of GCN filters are used. For the  $k$\textsuperscript{th} channel, the extracted user-rating vector is:
\begin{equation}
    \mathbf{U}^{(l,\mathcal{I}\rightarrow u, k)}=\sigma(\tilde{\mathbf{R}_k}\mathbf{I}^{(l)}\mathbf{W}^{(l,k)}_{R})
\end{equation}
Here, $\tilde{\mathbf{R}_k}$ is the column-normalized $\mathbf{R}_k$, and $\mathbf{I}^{(l)}$ is the item embedding of the $l$\textsuperscript{th} Rec-conv layer. Then each row of $\tilde{\mathbf{R}_k}\mathbf{I}^{(l)}$ represents the influence to one user $u$ from all items rated by $u$ with the rating value $k$.  $\mathbf{W}^{(l, k)}_{R} \in \mathbb{R}^{D\times D}$  is trainable. 

$K$ channels of outputs encode different users rating attitudes. 
% For example, $\mathbf{U}^{(l,i\rightarrow u, 1)}$, the output of the channel for rating 1, reflects users' dislikes and the one for the top rating $\mathbf{U}^{(l,i\rightarrow u, K)}$ describes users' most favors. 
Finally, we concatenate them and project to $D$-dim. to obtain final user-rating vectors:
% apply a one-layer neural network to reshape vectors:%~\citeauthor{ma2019disentangled}
\begin{equation}
    \mathbf{U}^{(l,\mathcal{I}\rightarrow u)}=\sigma(\Vert_{k=1}^{K}\mathbf{U}^{(l,\mathcal{I}\rightarrow u,k)}\mathbf{W}^{(l)}_{ur})
\end{equation}
where $\Vert$ is to concatenate matrices horizontally, and $\mathbf{W}^{(l)}_{ur} \in \mathbb{R}^{KD\times D}$ is a projector matrix.

\subsubsection{Attentive Aggregation For User Embedding}
There are three components to consider (i) $\mathbf{U}^{(l)}$ encodes user own factors from previous layer; (ii) social factor $\mathbf{U}^{(l,\mathcal{U}\rightarrow u)}$; and (iii) user-rating factor $\mathbf{U}^{(l,\mathcal{I}\rightarrow u)}$.

To differentiate impacts of these factors, we introduce attention mechanisms 
% $\mathbf{A}^{(l,t)} \in \mathbb{R}^{N\times D}, t\in \{u,u\rightarrow u,i\rightarrow u\}$, 
to control the information aggregated into the updated user embedding  $\mathbf{U}^{(l+1)}$ which is the input for the next Rec-conv layer. We let
$\mathbf{A}^{(l,u)}$ gate how much previous user embedding information to remember; $\mathbf{A}^{(l,\mathcal{U}\rightarrow u)}$ and $\mathbf{A}^{(l,\mathcal{I}\rightarrow u)}$ determine importance of the social factor and the user-rating factor. 

% Comparing to self-attention mechanism~\citep{gat}, where attention scores are singulars, we use attention vectors, which is more flexible to control how information is passed to theoutput~\citep{cho2014learning}.
Without loss of generality, we take $\mathbf{A}^{(l,u)}$ as an example. Calculation of $\mathbf{A}^{(l,\mathcal{U}\rightarrow u)}$ and $\mathbf{A}^{(l,\mathcal{I}\rightarrow u)}$ are similar. Formally, a single network layer (with parameter $\mathbf{W}^{(l,u)}_{A}$) is conducted to generate the contribution matrix: 
\begin{equation}
    \mathbf{A}^{(l,u)}=s([\mathbf{U}^{(l)} \Vert \mathbf{U}^{(l,\mathcal{U}\rightarrow u)} \Vert \mathbf{U}^{(l,\mathcal{I}\rightarrow u)}]\mathbf{W}^{(l,u)}_{A})
\end{equation}
where $s(x)=\frac{1}{1+e^{-x}}$ is the element-wise sigmoid operator. 

Then the updated user embedding $\mathbf{U}^{(l+1)}$ (i.e., inputs for the next Rec-conv layer) is obtained as the attention-weighted sum :
% of $\mathbf{U}^{(l)}$, $\mathbf{U}^{(l,u\rightarrow u)}$ and $\mathbf{U}^{(l,i\rightarrow u)}$ with their attention matrices:
\begin{equation}\label{eq:user-attention}
\begin{aligned}
    \mathbf{U}^{(l+1)} =& \mathbf{U}^{(l)}\odot\mathbf{A}^{(l,u)}+\mathbf{U}^{(l,\mathcal{U}\rightarrow u)}\odot\mathbf{A}^{(l,\mathcal{U}\rightarrow u)}\\
                        &+\mathbf{U}^{(l,\mathcal{I}\rightarrow u)}\odot\mathbf{A}^{(l,\mathcal{I}\rightarrow u)}
\end{aligned}
\end{equation}
where $\odot$ represents element-wise product.
% Note that the attention mechanism with shortcut from previous user latent factor $\mathbf{U}^{(l)}$ to its updated latent factor $\mathbf{U}^{(l+1)}$ alleviates the problem of over-smoothing in GNN~\citep{li2018deeper}. Thus, multiple Rec-conv layers can be stacked to capture more underlying information from social and user-item rating graphs. Please refer to Section \ref{sec:over-smoothing} for more details.

% \subsection{Disentangled GNN for aggregation}
\subsection{Update Item Embedding $\mathbf{I}^{(l)}\rightarrow\mathbf{I}^{(l+1)}$}\label{sec:item_emb}
To generate a latent vector for an item, not only its own previous item embedding but also the item-rated factor should be considered. We first generate item-rated vector, which encodes information from an item's overall rating. Then the attention mechanism is introduced.

\subsubsection{Item-Rated Vector $\mathbf{I}^{(l,\mathcal{U}\rightarrow i)}$}
The item-rated vector aggregates historical ratings of an item. Similar to the user-rating factor in Section \ref{sec:user-rateing-vector}, we adopt $K$ channels of GCN filters for diverse ratings. The output of the $k$\textsuperscript{th} channel is
\begin{equation}
    \mathbf{I}^{(l,\mathcal{U}\rightarrow i, k)}=\sigma(\tilde{\mathbf{R}_k^\intercal}\mathbf{U}^{(l)}\mathbf{W}^{(l,k)}_{RT})\\
\end{equation}
Here, $\tilde{\mathbf{R}_k^\intercal}$ is the column-normalized matrix of the transpose of $\mathbf{R}_k$. Then each row of $\tilde{\mathbf{R}_k^\intercal}\mathbf{U}^{(l)}$ collects effects on one item $i$ from all users who rate $i$ with the rating value $k$. $\mathbf{W}^{(l,k)}_{RT} \in \mathbb{R}^{D \times D}$ is the trainable transform matrix. 

Then we concatenate the $K$ outputs and connect a projection $\mathbf{W}^{(l)}_{ir} \in \mathbb{R}^{KD \times D}$ to get the item-rated vectors:
% With outputs from $K$ channels, we use a fully connected layer parametered by  $\mathbf{W}^{(l)}_{ir} \in \mathbb{R}^{KD \times D}$ to produce the item-rated vectors:
\begin{equation}
    \mathbf{I}^{(l,\mathcal{U}\rightarrow i)}=\sigma(\Vert_{k=1}^{K}\mathbf{I}^{(l,\mathcal{U}\rightarrow i,k)}\mathbf{W}^{(l)}_{ir})
\end{equation}

\subsubsection{Attention Aggregation For Item Embedding}
% Different items have various attributes, leading to diverse rating patterns. Empirically, users' ratings on virtual goods, such as songs and movies are relatively subjective. They evaluate these items mainly based on their tastes. In this case, when generating the item's latent vector, its own previous vector should dominate the ratings. On the other hand, ratings on a physical item, such as a personal computer, are highly relative to its quality, leading to relatively objective evaluations. To obtain the latent vector, the item-rated vector, which encodes ratings from users, should play a more important role.

To update the item embedding vectors $\mathbf{I}^{(l+1)}$, we introduce item attention mechanisms to assign contribution weights to the previous item vector $\mathbf{I}^{(l)}$ and the item-rated vector $\mathbf{I}^{(l,\mathcal{U}\rightarrow i)}$. Similar to the attention in user embedding updating, two item contribution matrices are:
\begin{align}
    \mathbf{B}^{(l,i)}&=s([\mathbf{I}^{(l)} \Vert \mathbf{I}^{(l,\mathcal{U}\rightarrow i)}]\mathbf{W}^{(l,i)}_{B})\\
    \mathbf{B}^{(l,\mathcal{U}\rightarrow i)}&=s([\mathbf{I}^{(l)} \Vert \mathbf{I}^{(l,\mathcal{U}\rightarrow i)}]\mathbf{W}^{(l,\mathcal{U}\rightarrow i)}_{B})
\end{align}
where, 
% $\mathbf{B}^{(l,i)}, \mathbf{B}^{(l,\mathcal{U}\rightarrow i)} \in \mathbb{R}^{M\times D}$ are attention matrices; 
$\mathbf{W}^{(l,i)}_{B} , \mathbf{W}^{(l,\mathcal{U}\rightarrow i)}_{B}\in \mathbb{R}^{2D\times D}$ are trainable parameter matrices. 

Then we obtain the updated item embedding $\mathbf{I}^{(l+1)}$ by: 
% by summing up $\mathbf{I}^{(l)}$ and $\mathbf{I}^{(l,\mathcal{U}\rightarrow i)}$ with attention weights:
\begin{equation}\label{eq:item-attention}
    \mathbf{I}^{(l+1)} = \mathbf{I}^{(l)}\odot\mathbf{B}^{(l,i)}+\mathbf{I}^{(l,\mathcal{U}\rightarrow i)}\odot\mathbf{B}^{(l,\mathcal{U}\rightarrow i)}
\end{equation}

% Users' purchase preferences are influenced by various factors, such as their own tastings and social effects from friends. To model these influences, in the $l$-th Rec-conv layer, we adopt graph convolutional networks to generate two kinds of user latent factors: user social factors $\mathbf{U}^{(l,u\rightarrow u)}$ and user rating factors $\mathbf{U}^{(0,i\rightarrow u)}$.

% Similarity, For items, item attention is used to differentiate influence from $\mathbf{I}^{(0)}$, item previous representations, and $\mathbf{I}^{(0,u\rightarrow i)}$, latent factors aggregating from user-item network, which encode items historical ratings. 

In summary, in each Rec-conv layer, using attention mechanisms, we dynamically aggregate social factor, user-rating factor and item-rated factor into the user/item embedding vector. 

% Note that the user and item embedding vectors are shared in the bipartite rating graph. This means that though there is no direct usage 

% Note that both user-rating vector and item-rated vector are extracted from the bipartite rating graph, these two factors are also mixed in user/item embedding updating process. In summary, Rec-conv layer effectively extracts intra-factor vectors and aggregates multiple factors with inter-factor attentions.  

We also emphasize that the attention mechanism 
% the gated-connection with previous user/item vectors (i.e., $\mathbf{U}^{(l)}$ and $\mathbf{A}^{(l,u)}$, and $\mathbf{I}^{(l)}$ and $\mathbf{B}^{(l,i)}$)
help to alleviates the over-smoothing issue in GNN~\citep{li2018deeper}. Thus, stacked Rec-conv layers can effectively capture user/item diverse information. Related evaluations are provided in Section \ref{sec:over-smoothing}.
\section{Experimental Study}
\label{sec:exp}
In this section, we perform extensive experiments to evaluate the performance of the proposed ASR method on two benchmarks, Ciao and Epinions. Social recommendation performances are shown in Section \ref{sec:exp-perf}. Section \ref{sec:diversity} further checks the diversity of predicted ratings. Model evaluations and ablation studies for ASR are followed in Section \ref{sec:analysis}. An efficiency evaluation is also provided in Section \ref{sec:efficiency}. 

\begin{table}
    \centering
    \begin{small}
    \caption{Statistics of datasets}
    \label{tab:dataset}
    \begin{tabular}{ccc}
    \hline
   Dataset &  Ciao &Epinions \\
   \hline
   \# Users ($|\mathcal{U}|$) & 7,317 & 37,311\\
   \# Items ($|\mathcal{I}|$) &104,975& 36,047\\ 
   \# Ratings ($|\mathcal{R}|$)& 111,781 & 1,054,202\\
   \# Social connections ($|\mathcal{S}|$)&283,319 & 428,034\\
    \bottomrule
    \end{tabular}
    \end{small}
    \vspace{-0.1in}
\end{table}

\subsection{Experimental Settings}\label{sec:settings}
\subsubsection{Datasets And Baselines}
Two benchmark datasets Ciao and Epinions\footnote{https://www.cse.msu.edu/~tangjili/datasetcode/truststudy.htm} are used. 
% \footnote{\red{We consider the social graphs (i.e., user-trust graphs) to be undirected in our experiments. This is consistent with settings applied in other works~\citep{wu2019dual,fan2019deep}.}}.
% These two datasets are extracted from consumer review websites Ciao (http://www.ciao.co.uk) and Epinions (www.epinions.com), respectively.
In these datasets, users can rate and give comments on items. The rating values are integers from 1 (like least) to 5 (like most). Besides, they can also select other users as their trusters. We use the trust graphs as social graphs. The statistics are summarized in Table~\ref{tab:dataset}.

% \subsubsection{Baseline methods.}
We compare ASR with state-of-the-art baselines with publicly available codes, including a traditional recommendation method (\umean, \imean, and \svd),  social recommendation methods (\soreg and \rste), and deep learning-based models (\graphrec, \danser). 
\umean and \imean predict an unobserved rating with the average of the user's and item's rating values, respectively.
\svd~\citep{koren2010factor} is a collaborative filtering method considering both explicit and implicit feedback. 
%\sorec is a probabilistic matrix factorization based method to solve the data sparsity and poor prediction accuracy problems. 
%\somf~\cite{jamali2010matrix} incorporates a mechanism of trust propagation in social graph to a matrix factorization model for the social recommendation.
\soreg~\citep{ma2011recommender} treats the social graph as regularization and use a matrix factorization framework. \rste~\citep{ma2009learning}  models users' favors and their friends' tastes with a probabilistic factor framework. \graphrec~\citep{fan2019graph} uses attention mechanisms to learn user/item embedding separately from social graph and user-item graph. \danser~\citep{wu2019dual} captures user/item dynamic/static features by GATs. 

{Note that we do not involve and compare with methods for top-$N$ recommendation, such as DiffNet~\citep{wu2019neural,wu2020diffnet}. Because top-$N$ recommendation targets to retrieve or rank $N$ items for users even utilizing temporal information instead of directly predicting user-item rating values. And the latter is our main purpose of social recommendation in this paper.}
% For \svd, \soreg, and \rste, we adopted the implementation from an open-source toolkit for the recommendation system\footnote{https://github.com/hongleizhang/RSAlgorithms}. 

\subsubsection{Parameter Settings}
For each dataset, we \textcolor{black}{randomly} split $80\%$ existing user-item ratings as the training set, $10\%$ as the validation set for tuning hyperparameters, and the left $10\%$ for testing. For ASR, we set the embedding dimension to 16, the batch size to 4096 and the learning rate to $0.0003$. We stack two Rec-conv layers (with GCN architecture\footnote{We evaluate effectiveness of different GNNs in extracting factors from the social graph and the rating graph in Section \ref{sec:disentabgled}.} and ReLU activation function) followed by an MLP for prediction. Early stopping is also used.
For other methods, we follow instructions in their papers to carefully tune hyperparameters, including but not limited to embedding size, batch size and learning rate, and report their best results.

\begin{table*}
\centering
% \begin{small}
\caption{Performance results on testing set}
\label{tab:ac}
\setlength\tabcolsep{15pt}
\begin{tabular}{c|ccc|ccc}
\hline
\multirow{2}{*}{Methods}   &\multicolumn{3}{c|}{Ciao} & \multicolumn{3}{c}{Epinions}\\
    \cline{2-7}
    & MAE$\downarrow$ & RMSE$\downarrow$ & P-ACC$\uparrow$ & MAE$\downarrow$ &RMSE$\downarrow$ & P-ACC$\uparrow$\\
    \hline
    \umean   &0.783 &1.033 &-- &0.951 &1.220 &--\\
    {\imean} &0.749 &1.016 &--&0.831 &1.081 &--\\
    \svd   &0.768 &1.043 &--&0.821 &1.096 &--\\
    \hline
    \rste  &0.753 &0.999 &0.548&0.828 &1.077 &0.714\\%0.715
    \soreg &0.746* &0.986* &0.558*&0.825* &1.064* &0.717*\\%0.729
    \hline
  \danser &0.740* &0.990* &0.630* &0.814* &1.076*& 0.722*\\ 
    \graphrec  &\textcolor{black}{0.745}* & \textcolor{black}{0.983}* &0.601*&0.820* &1.075*&0.719*\\
  \hline
  \our    &\textbf{0.717} &\textbf{0.962} &\textbf{0.637}&\textbf{0.781} &\textbf{1.048}&\textbf{0.730}\\

  \hline
  \multicolumn{7}{l}{$\downarrow$/$\uparrow$ next to metrics means that the smaller/larger, the better.}\\
  \multicolumn{7}{l}{* We do student-t test between ASR and the best baselines on each metric (* corresponds to $p<0.05$).}\\
\end{tabular}
% \vspace{-0.1in}
% \end{small}
\end{table*}

\subsection{Performance Comparison}\label{sec:exp-perf}
We adopt two widely used metrics, mean absolute error (MAE) and root mean square error (RMSE)~\citep{wu2019dual}. Smaller MAE and RMSE scores indicate better performance. We repeat each experiment 5 times and report average results on testing set in Table~\ref{tab:ac} (first two columns of each dataset). The best performer is highlighted with bold fonts. Note that small improvements in MAE or RMSE will lead to significant enhancement on the performance of the top-$N$ recommendation~\citep{fan2019graph}. From the table, we observe:
\begin{itemize}
     \setlength\itemsep{0.15em}
    \item Among the matrix factorization-based methods (\svd, \soreg, and \rste), \svd only utilizes user-item rating information, while the two better performers, \soreg and \rste, use the social graph as additional information.  Comparing them verifies that social factor can provide complementary information for recommendations. 
    
    Interestingly, we observe that in some cases, the simple method \imean outperforms \svd in Ciao and Epinions (RMSE). This observation is similar as that pointed out by~\citet{dacrema2019we}. They also found that simple methods
    % like TopPopular, and ItemKNN 
    may achieve comparable performances with more complicated alternatives. 
    
    \item In general, GNN-based methods, including \our, \graphrec, and \danser, outperform traditional social recommendation baselines. Because GNN models can more effectively aggregate information from both social and rating graphs. The comparison between these two types of methods reflects the power of GNN for social recommendation systems. 
    
    \item \our outperforms others in both datasets with statistical significance. 
    % Specifically, ASR relatively improve MAE and RSME in Ciao over the second-best baseline 2.98\% and 0.92\%, respectively.  The improvement on Epinions is 4.05\% and 1.50\% respectively. 
    This is because \our considers the diversity of users and items and combine multiple diverse factors. In \our, attention mechanisms can actively extract and aggregate the social, user-rating and item-rated factors from the social and rating graph. The disentangling strategy also differentiates impacts of different rating values. Detailed evaluations are shown in Section \ref{sec:analysis}. 
    % These improvements in both metrics verify the superiority of ASR.
\end{itemize}

\subsection{Diversity of User-Item Ratings}\label{sec:diversity}
Except for ASR's better results on MAE and RMSE, in this section, we demonstrate that ASR can obtain more accurate and diverse ratings than baselines.  

Because MAE and RMSE are both for the overall performance, we now check whether a model can predict accurate ratings for each user with the \textit{pairwise-ranking accuracy (P-ACC)}. Suppose that one user has distinct ratings for two items, we define \textit{a hit} when one algorithm can predict the ratings with the correct relative ranking. Then P-ACC reflects the overall hitting rate. Higher P-ACC means better performance.  Table~\ref{tab:ac} (the third column of each dataset) shows the P-ACC results over the testing set. \textcolor{black}{Note that we only consider the top-5 performers}. Results evidence that ASR can predict more accurate individual rating-rank. 

\begin{figure}[t]
    \centering
    \includegraphics[width=1.7in]{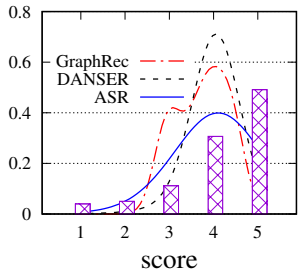}
    % \hspace{-1.3em}
    % \includegraphics[width=1.7in]{Figure/Exp/epinion_distribution2.eps}
    \vspace{-0.1in}
    \caption{Distributions of predicted ratings in Ciao test set}
    \label{fig:score_diversity}
    \vspace{-0.1in}
\end{figure}

Next, we check the goodness of overall rating distributions of ASR and the other two best baselines \danser and \graphrec. The bars in Figure \ref{fig:score_diversity} are the ground-truth rating histogram of Ciao testing set. We can see that \graphrec predicts most ratings as ``3'' and ``4'' and neglects other values even for the most value ``5'' in the ground-truth. \danser even narrows its all ratings around ``4'' (values in [3.6, 4.3] covers more than 80\% of its ratings). However, \our can fit better with the ground-truth by considering the rating diversity.

\subsection{Model Analysis}\label{sec:analysis}
In this section, we conduct ablation study for \our. Three aspects are evaluated: the necessity of the two main components of \our, i.e., inter-factor attention mechanisms and the disentangling strategy; and the sensitivity of \our to over-smoothing of GNNs~\citep{li2018deeper}. 

\subsubsection{Impact of Inter-Factor Attention Mechanisms}
 \label{sec:exp:att}
To capture inter-factor contribution, we apply attentions in user/item embedding updating. To show the effectiveness of the attention, we compare \our with two variants:
\begin{itemize}
    \item \ouru removes the user attention, i.e., $\mathbf{A}^{(l,u)}, \mathbf{A}^{(l,\mathcal{U}\rightarrow u)}$ and $\mathbf{A}^{(l,\mathcal{I}\rightarrow u)}$ in Equation \eqref{eq:user-attention} are set with all 1s. 
    \item \ouri removes the item attention. i.e., $\mathbf{B}^{(l,i)}$ and $\mathbf{B}^{(l,\mathcal{U}\rightarrow i)}$ in Equation \eqref{eq:item-attention} are set with all 1s.  
\end{itemize}

\begin{figure}[!t]
    \centering
    \subfigure[MAE]{\includegraphics[width=1.75in]{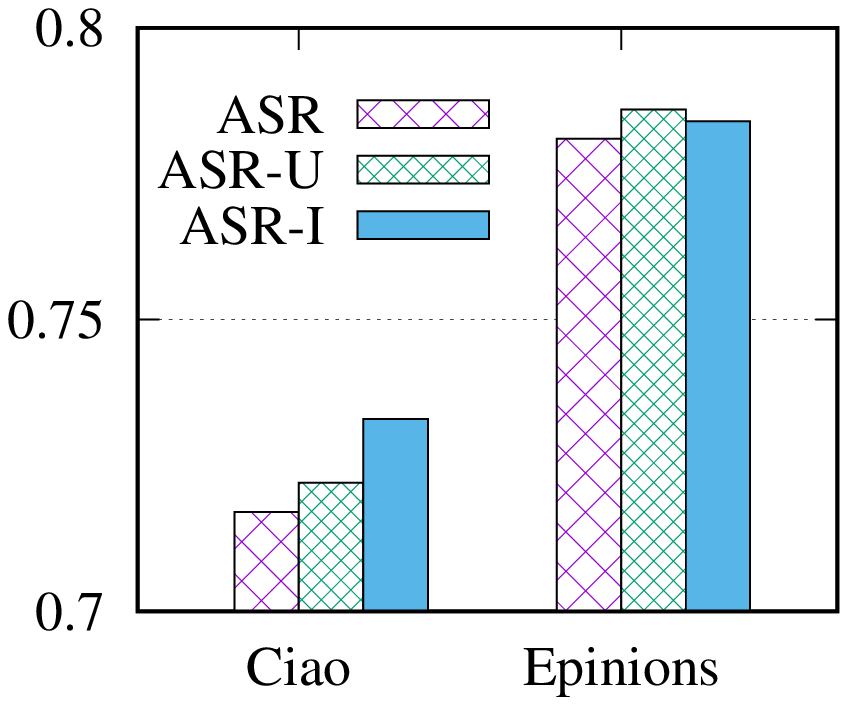}\label{fig:att:mae}}\hspace{-1.8em}
    \subfigure[RMSE]{\includegraphics[width=1.75in]{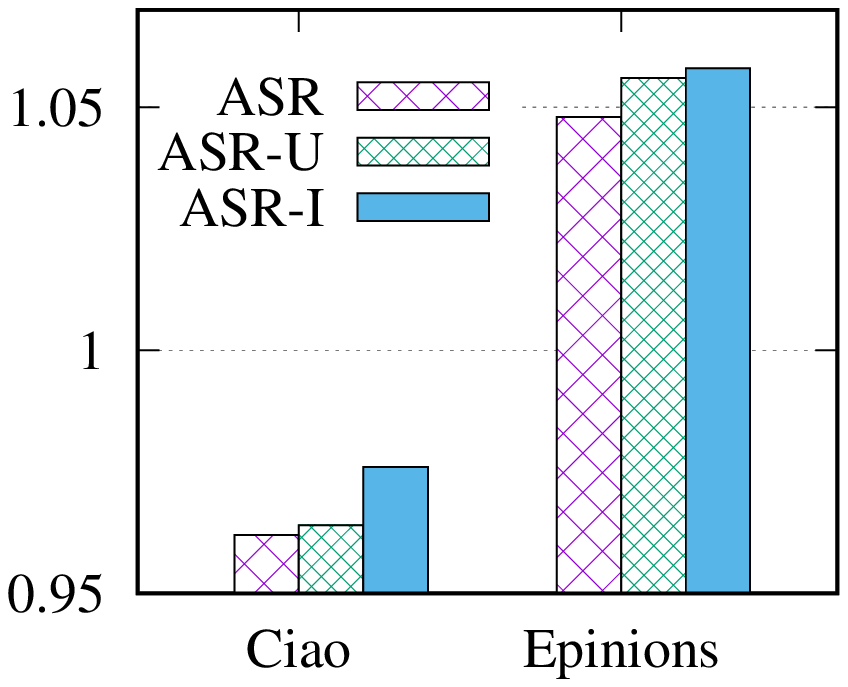}\label{fig:att:rmse}}
    \vspace{-0.1in}
    \caption{Impact of user/item attention mechanism}
    \label{fig:ablation1}
    \vspace{-0.1in}
\end{figure}

MAE and RMSE results of ASR and its two variants are shown in Figure~\ref{fig:ablation1}. We can see that \our achieves the best MAE and RMSE scores in both Ciao and Epinions. Comparing \our with \ouru and \ouri, we conclude that including user/item inter-factor attentions provides \our powerful and flexible abilities to effectively aggregate impacts from multiple factors, which leads to better results. 
% This demonstrates the necessity of attention mechanisms.

\subsubsection{Impact of The Disentangling Strategy And GNNs}\label{sec:disentabgled}
The disentangling strategy of \our first splits the entire rating graph into subgraphs based on distinct rating values, and then aggregates GNN-extracted features from each subgraph. 
% In contrast, the traditional graph convectional network ignore the edge attributes. 
To verify its advantage, we compare \our with its variant: \ourgatr which ignores diverse rating values and adopts GAT on the entire rating graph $G_R$. In \ourgatr, though rating value differences are removed in $G_R$, GAT can assign distinct weights on different items for one user or different users for one item. Note that GAT is also used in both the two most competitive baselines \danser and \graphrec. 

Comparing the first two rows in Table \ref{tab:other_gnn}, \our outperforms \ourgatr significantly. Although the GAT in \ourgatr aggregates neighbors' features with different weights in the rating graph, it cannot effectively leverage the diverse rating scores. However, the disentangling strategy in \our guarantees that diverse propagating patterns of diverse ratings can be distinguished.

To check the effect of different GNN, in \our, we also replace the GCN which is applied in the social graph $G_S$ with a GAT but keep the disentangling strategy in the rating graph $G_R$ (\ourgats). Comparing \our with \ourgats in Table \ref{tab:other_gnn}, we see that applying GAT in the social graph $G_S$ can obtain comparable performances as the original ASR. 

Comparing the three together, we conclude that the disentangling strategy in the rating graph plays an essential role in ASR. Both GAT and GCN can be applied in the social graph to effectively extract the user-social factor.

\begin{table}
\centering
\begin{small}
\caption{ASR v.s. its variants}
\label{tab:other_gnn}
\begin{tabular}{c|cc|cc}
\hline
\multirow{2}{*}{Methods}   &\multicolumn{2}{c|}{Ciao} & \multicolumn{2}{c}{Epinions}\\
    \cline{2-5}
    &MAE$\downarrow$ & RMSE$\downarrow$  &MAE$\downarrow$ &RMSE$\downarrow$ \\
    \hline
    ASR &0.717 & 0.962  &  0.781 &1.048  \\
    ASR-w/-$G_R$-GAT & 0.789 & 1.027 &0.803   &1.076\\
    ASR-w/-$G_S$-GAT & 0.720 & 0.971 & 0.793  &1.058 \\ 

  \hline
\end{tabular}
\end{small}
\end{table}

\begin{figure}[!t]
    \centering
    \hspace{-0.15in}
    \subfigure[MAE]{\includegraphics[width=1.22in]{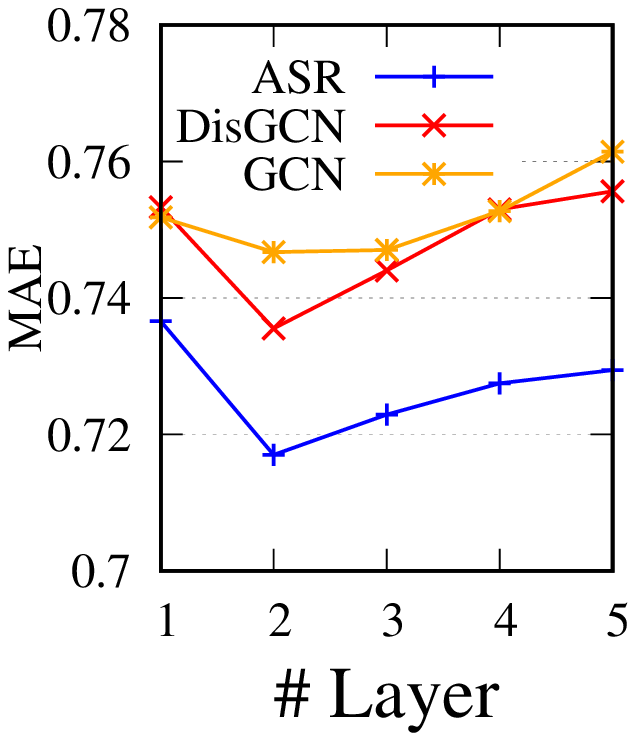}\label{fig:l:mae}}\hspace{-0.17in}
    \subfigure[RMSE]{\includegraphics[width=1.22in]{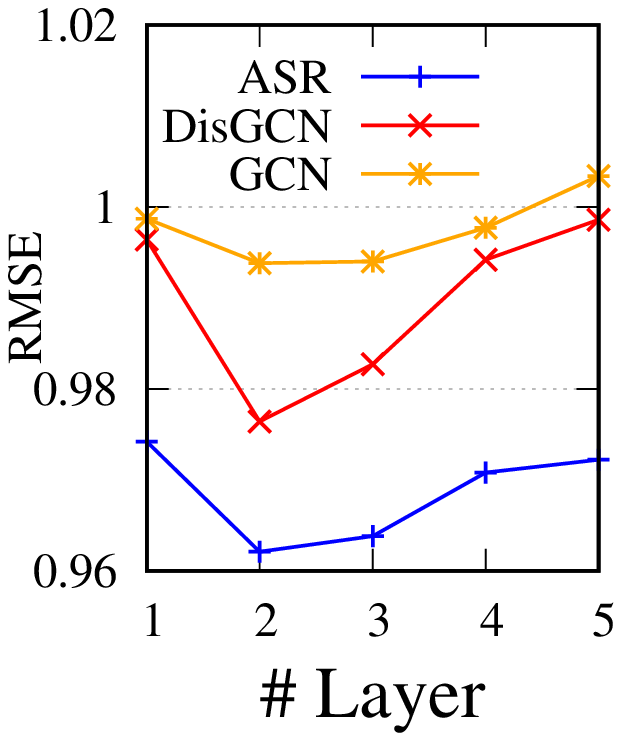}\label{fig:l:rmse}}\hspace{-0.17in}
    \subfigure[{Distance}]{\includegraphics[width=1.22in]{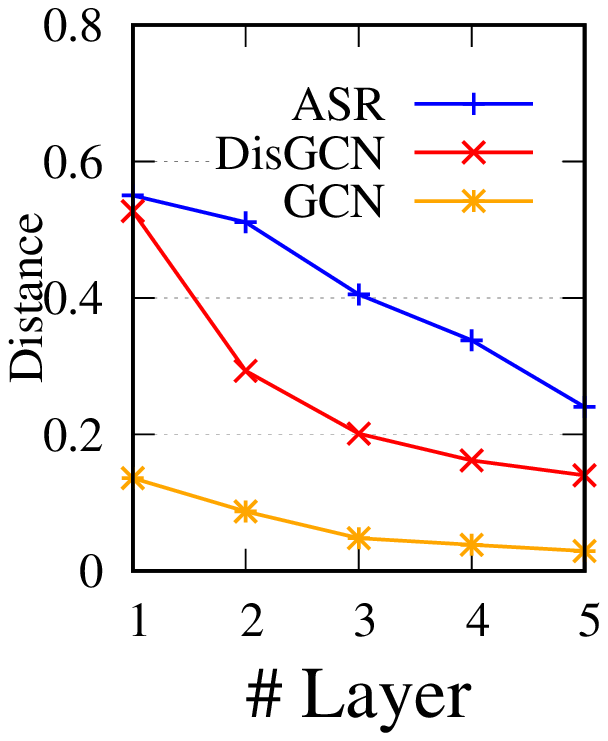}\label{fig:l:distance}}
        \vspace{-0.1in}
    \caption{Impacts of stacking Rec-conv layers on Ciao dataset}
    \label{fig:para:l}
        \vspace{-0.1in}
\end{figure}

\subsubsection{Stacking Rec-conv Layers V.S. Over-smoothing}\label{sec:over-smoothing}
With more GNN layers stacked, most GNNs suffer from over-smoothing issue~\citep{li2018deeper}, which means that embedding vectors tend to be similar. If we compute the pairwise-distance (e.g., cosine distance) of the embedding vectors, then the more layers are, the smaller the average distance is. The larger the distance is, the less the over-smoothing is. To evaluate the sensitivity of \our to the over-smoothing, we compare \our with its two GNN variants (i) DisGCN: removing the attentions in \our; and (ii) GCN: removing attentions and disentangling in \our. Results over different numbers of stacked layers are shown in Figure~\ref{fig:para:l}. The results demonstrates that \our can alleviate the over-smoothing more than the two variants. 

ASR can alleviate the over-smoothing for two reasons. First, comparing \our with DisGCN, user/item attentions can actively control how much impacts from different factors to pass forward to the next layer. Besides, comparing GCN with DisGCN and \our, the disentangling strategy splits the mixed ratings in the rating graph and diversifies effects of different ratings in the user/item embedding updating processes. As a result, we can stack multiple Rec-conv layers when encountering complex datasets to aggregate more underlying factor features.

\subsection{Efficiency Evaluation}\label{sec:efficiency}
In \our, GCNs are applied for distinct rating values. This brings in more parameters than when considering the rating graph as a whole graph. However, in practice, we found that \our executes very fast. We compare the training time per epoch of \our and the two deep learning based baselines \danser and \graphrec. Note that we use the same embedding dimension for the three models. As shown in Table \ref{tab:efficiency}, \our is the most efficient model. We analyze reasons as the following. Multiple GATs are used in both \danser and \graphrec. Different with GCN, attention weights computation in GAT costs quadratic time in terms of number of nodes. Besides, a policy network exists in \danser to search optimal returns. In \graphrec, the current implementation\footnote{https://github.com/wenqifan03/GraphRec-WWW19} takes each user-item rating pair as a minibatch, because it specifies individual user who may have different number of user-neighbors and rated items.  This means that trainable parameters will be updated for each rating pair which is time-consuming.  

%\our because we use GCN instead of GAT. In GAT, attention computation costs  quadratic time in terms of number of nodes.

\begin{table}
\centering
\begin{small}
\caption{Training time (sec) per epoch}
\label{tab:efficiency}
\begin{tabular}{cccc}
\hline
Method & Ciao & Epinion\\ \hline
\danser    & 76.372 & 272.78 \\
\graphrec  & 862.10 &1786.37 \\
\our & 15.53 & 28.90 \\
  \hline
\end{tabular}
\end{small}
\end{table}

\section{Conclusion}
\label{sec:conclusion}
% In this paper, we proposed an attentive social recommendation system \our. \our can effectively capture diversity of users and items with intra-factor and inter-factor features extracted from the social and rating graphs. The proposed Rec-conv network layer including attention mechanisms and the disentangling strategy enables \our to attentively aggregate multiple factors when embedding user/item representations. The disentangling strategy also helps \our diversify effects of different ratings. Comprehensive experiments on two benchmark datasets demonstrate the effectiveness of \our over state-of-the-art baselines. Ablation studies verified the necessity of \our components.

In this paper, we proposed an attentive social recommendation method \our. \our includes user and item attentions to capture diversities of users and items. The proposed Rec-conv network layer and attention mechanisms enable \our  to actively extract and fuse the social factor, user-rating factor and item-rated factor from user social graph and user-item rating graph. In addition, a disentangling strategy was developed to aggregate information from diverse ratings in the user-item rating graph. Comprehensive experiments on two benchmark datasets demonstrate the effectiveness of \our. Ablation studies on attention mechanisms, disentangling strategy and GNNs verified the necessity of these modules. We can stack multiple Rec-conv layers in \our but less being affected by the over-smoothing. The training process of \our is much faster than alternatives as well. 
% The future work involves considering social recommendation with user/item attributes and designing attention mechanisms inside the social and user-item rating graph.

% For example, in Epinions dataset, reviews from users are available, which can be considered as item attributes.
% Second, attention mechanisms design inside social and user-item rating graphs is also in process. 

% when learning user/item representations. In addition, a new disentangled GCN was developed to aggregate information from neighborhoods in user-item rating graph. Comprehensive experiments on two benchmark datasets demonstrate the effectiveness of \our. Ablation studies on attention mechanisms and disentangled GCN verified the necessity of these modules. A couple of topics can be further investigated. First, we would like to consider social recommendations with user/item attributes. For example, in Epinions dataset, reviews from users are available, which can be considered as item attributes. Second, attention mechanisms design inside social and user-item rating graphs is also in process. 

\clearpage
\bibliography{reference}
\end{document}